\documentclass{article}
\usepackage{neurips_2020, hyperref, amsfonts, amsmath, graphicx, lineno, xcolor, setspace}

\title{Conditioning on PDE Parameters to Generalise Deep Learning Emulation of Stochastic and Chaotic Dynamics}

\author{
Ira J. S. Shokar$^1$, Rich R. Kerswell$^1$, Peter H. Haynes$^1$   \\
$^1$Department of Applied Mathematics and Theoretical Physics, University of Cambridge,\\ Wilberforce Road, Cambridge, CB3 0WA, UK\\
}

\begin{document}
\begin{spacing}{1.15}
\maketitle


\begin{abstract}
We present a deep learning emulator for stochastic and chaotic spatio-temporal systems, explicitly conditioned on the parameter values of the underlying partial differential equations (PDEs). Our approach involves pre-training the model on a single parameter domain, followed by fine-tuning on a smaller, yet diverse dataset, enabling generalisation across a broad range of parameter values. By incorporating local attention mechanisms, the network is capable of handling varying domain sizes and resolutions. This enables computationally efficient pre-training on smaller domains while requiring only a small additional dataset to learn how to generalise to larger domain sizes. We demonstrate the model's capabilities on the chaotic Kuramoto-Sivashinsky equation and stochastically-forced beta-plane turbulence, showcasing its ability to capture phenomena at interpolated parameter values. The emulator provides significant computational speed-ups over conventional numerical integration, facilitating efficient exploration of parameter space, while a probabilistic variant of the emulator provides uncertainty quantification, allowing for the statistical study of rare events.
\end{abstract}


\section{Introduction}
\label{chap:EDS_intro}

Recent investigations have examined cost-effective alternatives to physics-based solvers, given the computational demands of modelling high-degree-of-freedom systems, such as turbulent flows. Machine learning has emerged as a promising direction \cite{fluids_ML_Ricardo_Brunton}. However, most machine learning methods encounter challenges in accurately generalising beyond their training distribution. This issue is particularly salient in the emulation of dynamical systems, where the physical parameters governing the partial differential equations (PDEs) may differ from those encountered during training, potentially affecting the model's ability to generalise across parameter regimes. One approach to address this is transfer learning \cite{transfer}. This involves freezing the learned weights of a base-trained network and training additional layers to generalise to new tasks \cite{lineartransfer, deeponet, transfernonlinear}. Alternatively, the transformer architecture \cite{Attention} has demonstrated remarkable performance in generalisation tasks without the need for additional learnable parameters, which can be attributed to its attention mechanism for self-supervised learning during pre-training \cite{BERT}.

We pre-train a transformer-based network on a dataset comprising trajectories generated from a single parameter choice for a PDE to enable the neural network to initially focus on learning dynamical structures independent of variations across different parameter regimes. Subsequently, we fine-tune the model on a smaller dataset encompassing trajectories from a range of parameter values, which facilitates generalisation across different regimes \cite{iclr}. The transformer architecture incorporates two key modifications to the standard implementation \cite{Attention}: (i) local attention \cite{SASA}, which captures spatial interactions within the neighbourhood of each point, and (ii) adaptive layer normalisation \cite{FILM}, which enables the network to learn transformations of latent variables that are conditioned on varying parameter values.

In fluid dynamics, spatial correlation lengths are typically finite, a property that can be exploited to make computation more efficient by restricting message passing to local spatial regions. Several neural network architectures incorporate this locality inductive bias, with convolutional neural networks (CNNs) \cite{cnn} being among the most prominent. Convolutional layers learn filters that operate over localised image patches, enabling weighted local aggregation. However, their fixed weights limit adaptability across different spatial scales and input fields of varying sizes. To address this limitation, we employ local attention \cite{SASA}, which replaces the static weights in convolutional kernels with adaptive weights based on attention mechanisms. This flexibility enables the neural network emulator to handle inputs of varying sizes by adaptively capturing spatial correlations across scales. Through dynamic local attention, the model learns local interactions, accommodating different domain sizes and scales.

We contrast this approach with operator learning methods, such as the Fourier Neural Operator (FNO) \cite{FNO}. The FNO exhibits a low spatial frequency bias, which can limit its ability to capture fine-scale interactions crucial for accurately modelling larger-scale dynamics. It also affects systems with a high number of dominant modes, where the model's capacity to capture essential dynamics across scales is further constrained. Additionally, the FNO requires a fixed number of wavenumbers to represent the dynamics, which we show restricts its generalisability to systems with significant variation in domain size and complexity.

There is a growing body of research aimed at developing generalisable neural network surrogates with the advent of foundation models, neural networks trained on diverse datasets that can adapt to multiple tasks, offering significant promise in this area. 
Recent work training across multiple systems \cite{MPP} highlights the potential of providing better-informed, physics-based priors. While foundation models show significant promise for generalising across families of physical systems, not merely parameter spaces, their practical exploration presents significant challenges. The computational demands are substantial, requiring vast amounts of diverse training data and extensive computing infrastructure. Given these practical constraints and the focused scope of our investigation into specific systems, we have opted to pursue more a targeted approach, leaving the exploration of foundation models as a promising direction for future research.

To develop a neural network emulator capable of being conditioned on the parameters of the PDE, we implement adaptive layer normalisation \cite{FILM}. This approach involves replacing the affine transformations in layer normalisation \cite{layernorm} with learned functions of the conditioning inputs, in our case PDE parameters. As a result, the model can adaptively scale and shift the hidden states within the network based on contextual information, modifying the network's behaviour according to the parameter values.

We illustrate that our approach can be pre-trained on smaller input fields and subsequently fine-tuned on larger ones, allowing for inference on systems with diverse numbers of degrees of freedom. We demonstrate our approach using two PDEs: the Kuramoto-Sivashinsky (KS) equation, a well-studied chaotic PDE, and beta-plane turbulence, a well-studied stochastically-driven PDE. The neural network serves as a resource-efficient emulator of the systems, providing autoregressive roll-outs that remain indefinitely stable while effectively capturing the statistical properties across parameter regimes. Furthermore, we demonstrate that a probabilistic variant of the emulator enables the efficient generation of large ensembles, allowing for the quantification of extreme event likelihoods in a computationally tractable manner.

\section{Methodology}
\label{chap:EDS_methodology}

\subsection{Neural Network}
\label{chap:EDS_PT}

\begin{figure}[t!]
\begin{center}
\includegraphics[width=\textwidth]{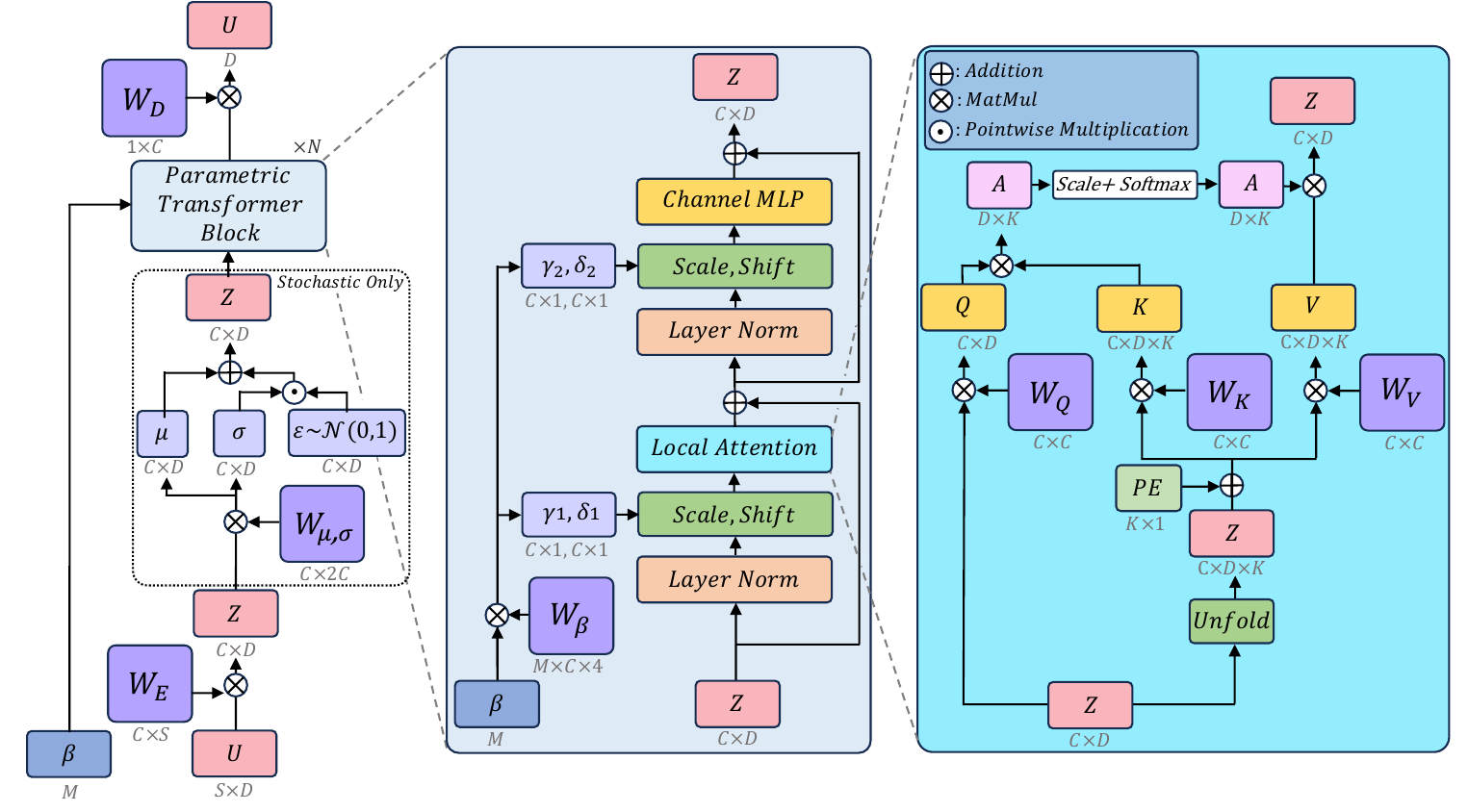}
\end{center}
\caption{Schematic of the neural network architecture. The network is structured around a transformer architecture conditioned on parameter $\beta$ (or $L$ in the case of the KS equation) and initialised by conditions $U$. For probabilistic outputs, a sampling operation is applied before the $N$ transformer blocks. Within each transformer block, adaptive layer normalisation conditions the transformer on $\beta$ by replacing scale and shift parameters. In each local attention block, an unfold operation is performed on keys (K) and values (V) to stack neighbouring pixels within the specified window size, $K$, followed by the application of learned relative positional encodings (PE). Here, Q denotes query vectors, and A represents the attention matrix - although in practice, this is not materialised through the implementation of Flash Attention \cite{FlashAttention}. Each $W$ denotes learned weights for linear transformations, with arrows indicating the forward pass. In this study, the conditioning parameter $\beta$ is defined with a size of $M=1$. However, the architecture is flexible to extension to larger dimensions within the parameter space}
\label{fig:ml_arch}
\end{figure}

\subsubsection{Local Attention}
\label{chap:EDS_LA}

Partial Differential Equations (PDEs) are often solved over varying domain sizes, and ideally, a neural network emulator should adapt to these varying input sizes and scales. To address this challenge, we employ local attention, as introduced in \cite{SASA}. This technique focuses attention on a localized window around each spatial point, similar to the receptive field in convolutional networks. This approach enables the model to efficiently capture essential local spatial dependencies while modelling interactions across multiple scales hierarchically through transformer layers. Local attention learns fine-scale interactions, while global structures are progressively captured across layers. By using attention mechanisms instead of convolutional operations, this method dynamically computes spatial weights, maintaining flexibility for different input scales. This is in contrast to convolution kernels, which learn fixed spatial correlations and thus lack the adaptability to accommodate varying input scales.

To transform the inputs, $u_t \in \mathbb{R}^{D \times S}$, into a latent space $z \in \mathbb{R}^{D \times C}$ within which the transformer operates, we perform a temporal uplift \cite{FNO}, where $S$ is the temporal history sequence length (for full resolved fields $S=1$, while $S=2$ for under-resolved systems, which we further detail in Section \ref{chap:EDS_prob}) and $D$ is the number of grid points in space. This encoding transformation takes the form of a linear transformation defined by weights $W_{\text{E}} \in \mathbb{R}^{S \times C}$, where $C$ represents the number of channels. After passing through the transformer blocks, a decoder linear transformation maps from channels $C$ to a single channel, yielding the prediction $\tilde{u}_{t+1} \in \mathbb{R}^D$. Figure \ref{fig:ml_arch} provides a schematic of this architecture.

The local attention layer is designed to compute correlations within local neighbourhoods centred around each position in the latent space. For a specific position $x$ in $z \in \mathbb{R}^{C \times D}$, we define the local neighbourhood $\mathit{N}_K(x)$ with a spatial extent of $K \in \mathbb{Z}^{+}$. The attention mechanism computes a weighted sum of the values in this neighbourhood, expressed as $a_x = \sum_{b \in \mathit{N}_k(x)} \text{softmax}_b(q_x^T k_b) v_b$ where $q \in \mathbb{R}^{C \times D}$, $k \in \mathbb{R}^{C \times D}$, and $v \in \mathbb{R}^{C \times D}$ represent the query, key, and value vectors, respectively, that are each obtained through linear projections of $z$ \cite{Attention}. Specifically, $q_x$ corresponds to the query vector for position $x$, while $k_b$ and $v_b$ are the key and value vectors associated with neighbouring positions $b \in \mathit{N}_K(x)$. To implement local attention we employ the unfold operation. This operation rearranges the input tensor $z$ into overlapping local windows of size $K$, \( \mathbb{R}^{C \times D} \rightarrow \mathbb{R}^{C \times D \times K} \). In our implementation, we also employ circular padding to ensure that the neighbourhoods wrap around at the boundaries of the input tensor, to retain translational equivariance.

Local attention enables the use of large kernel sizes in attention mechanisms without introducing additional parameters for the kernel size, as this is managed by the linear projection layers. The advantage of local attention is that it significantly reduces both computational load and memory usage from $D^2$ in full attention to $D \times K$, where $D > K$. To encode relative positional information between elements, we introduce a learnable relative positional encoding, $\text{PE} \in \mathbb{R}^{K}$. This encoding contributes only an additional $K$ parameters, whereas convolutional operations scale the parameter count by $C_{\text{in}} \times C_{\text{out}} \times K$, where $C_{\text{in}}$ and $C_{\text{out}}$ are the input and output channel dimensions, respectively. Consequently, we can achieve a large effective receptive field with fewer layers, eliminating the need for pooling or striding. This capability is particularly advantageous in systems with multi-scale dynamics. By focusing on relative positions rather than absolute ones, our attention mechanism, alongside the implementation of circular padding in the unfold operation, maintains translation equivariance, thus helping the model respect the symmetries inherent to the dynamics.

\subsubsection{Parametric Transformer}
\label{chap:EDS_FILM}

Unlike generative models in the language and image domains that rely on discrete text for conditioning—necessitating encoding into the model's latent space—our model is conditioned on continuous scalar variables (PDE parameters) which do not require any prior transformation. We explored various approaches for conditioning on parameter information, which include: (i) introducing conditioning values as features to the linear transformations in each attention block, (ii) concatenating conditioning information with fully-connected layers and (iii) introducing conditional values through cross attention \cite{nPDE}. In this work, we replace feature-wise affine transformation parameters in layer normalisation with a learned function of the conditioning information, as described in \cite{FILM}. While all methods yielded comparable results, our final approach facilitated more efficient pre-training by decoupling the scale and shift parameters from the conditioning variables, allowing for more effective transformations during fine-tuning - see further details in Section \ref{chap:training}.

We define a function that generates scaling parameters, $\gamma$, and shift parameter, $\delta$, to modulate the intermediate activations for both the attention operation and the multi-layer perceptron (MLP), totalling 4. These parameters are obtained via linear transformation $W_\beta \in \mathbb{R}^{M \times 4C}$ applied to the conditioning inputs for each transformer block, which we denote as $\beta$, where $M=1$ represents the size of the context vectors $\beta$. Although this work focuses on the case of $M=1$, the framework is designed to be flexible, allowing for straightforward extension to larger dimensions within the parameter space. The transformations within each transformer block are formulated as follows:

\begin{align}
    z &\rightarrow z + \mathrm{LA}\left(\gamma_1 \cdot \mathrm{LN}(z) + \delta_1\right), \\
    z &\rightarrow z + \mathrm{MLP}\left(\gamma_2 \cdot \mathrm{LN}(z) + \delta_2\right).
\end{align}

Here, $z$ represents the hidden state of the transformer block, $\mathrm{LA}$ denotes the local attention mechanism, $\mathrm{LN}$ indicates layer normalisation, and $\mathrm{MLP}$ refers to the multi-layer perceptron component, comprising two linear transformations, with an intermediate GELU \cite{gelu} activation.

The effectiveness of the adaptive normalisation scheme arises from its ability to modify the representational capacity of the model without introducing explicit multiplicative interactions between the conditioning inputs and hidden activations. Instead, the external dependencies are incorporated via learned affine transformations, ensuring stability during training and preserving the transformer’s ability to learn hierarchical representations. 

\subsubsection{Probabilistic Modelling}
\label{chap:EDS_prob}

The emulator is also designed to model under-resolved dynamics—systems where the input fields have fewer degrees of freedom than required for explicitly solving the underlying PDE. To account for the uncertainty introduced by these unresolved dynamics, probabilistic forecasting is often necessary \cite{Stochasticparametrization}. To introduce stochasticity into the neural network, we incorporate a sampling layer between the encoder and the transformer blocks. Given the encoded latent representation $z$, we define the parameters $\mu(z)$ and $\sigma(z)$ through linear projections, specifying a Gaussian distribution $\mathcal{N}(\mu(z), \sigma^2(z))$ from which latent samples are drawn, following the approach introduced in the variational autoencoder (VAE) \cite{VAE}.  

As in the VAE, we apply the reparametrisation trick to enable backpropagation: a noise term $\epsilon \sim \mathcal{N}(0, I)$ perturbs the sampled latents via $z := \mu(z) + \sigma(z) \cdot \epsilon$. Here, $\sigma(z)$ acts as a state-dependent scaling factor for the stochastic noise $\epsilon$, conditioned on the input $z$. Sampling $\epsilon$ enables ensemble generation with different noise realisations—analogous to sampling $\zeta$ in the beta-plane model. The transformer blocks then learn a mapping from the latent distribution $\mathcal{N}(\mu(z), \sigma^2(z))$ to the target data distribution.

\subsection{Objective Function}
\label{chap:CRPS}

In deterministic prediction within an operator learning framework—where the emulator receives the full state information at time $t$ along with a single history step ($S=1$)—the neural network is pre-trained and fine-tuned using the mean squared error (MSE) loss, following standard practice. The loss function is defined as $\mathcal{L} = MSE(U, \tilde{U})$, where $U$ denotes the ground truth trajectory from the training dataset, and $\tilde{U}$ represents the emulator's forecast.

However, when training the emulator to make probabilistic forecasts in the under-resolved settings, the neural network is pre-trained and fine-tuned using the following loss function, which combines the Continuous Ranked Probability Score (CRPS) \cite{CRPS}, a proper scoring rule that generalises the Mean Absolute Error (MAE), with a spectral loss to ensure energy preservation across all scales:

\begin{equation}
\mathcal{L} = \mathrm{CRPS}\left(U_{t+1}, \tilde{U}_{t+1}\right) + \lambda \mathrm{MAE}\left(|\mathcal{F}[U_{t}]|, |\mathcal{F}[\tilde{U}_{t}]|\right).
\label{eq:loss}
\end{equation}

The CRPS is defined as follows:

\[
\mathrm{CRPS}(U, \tilde{U})= \frac{1}{m} \sum_{i=1}^m\left| U-\tilde{U}^{(i)}\right| - \frac{1}{2m^2} \sum_{i=1}^m \sum_{j=1}^m\left|\tilde{U}^{(i)}-\tilde{U}^{(j)}\right|
\]

where $U$ represents the truth trajectory from the training dataset and $\tilde{U}^{(i)}$ is the $i^{\text{th}}$ member of the prediction ensemble, with the model producing an ensemble of size $m$ for time step $t+1$, and $\mathcal{F}$ is the discrete Fourier transform. The network is trained to minimise the difference between each predicted ensemble member and the truth trajectory $U_{t+1}$ - a metric to determine forecast skill, while simultaneously maximising the dissimilarity among individual ensemble members $\tilde{U}^{(i)}_{t+1}$ due to the inclusion of the second term in the CRPS - a metric of forecast spread. The CRPS encourages ensembles that exhibit a spread that is approximately equivalent to the forecast skill, which is a common measure of quality for probabilistic forecasts. Although performing a forward pass over an ensemble increases training costs, empirical results indicate that using an ensemble size of $m=2$ is sufficient, as was also observed in \cite{neuralgcm}.

Alternative approaches to generative modelling include diffusion models \cite{diffusion}, which also aim to map from a latent distribution to a target distribution. However, they define a stochastic process characterised by an ordinary differential equation (ODE) that transforms a normal distribution into a target distribution while conditioning on previous time steps through mechanisms like cross-attention. Diffusion models, in particular, utilise a sequence of de-noising steps, requiring multiple forward passes to generate a single output. While our approach utilising the CRPS requires $m=2$ passes through the network to produce an ensemble forecast with 2 members to estimate the CRPS, this is significantly fewer than the 20 to 50 steps typically needed by diffusion models. Moreover, these forward passes can be executed in parallel, provided there is adequate memory overhead, which offers a clear advantage over the sequential nature of diffusion models which precludes such parallelism. As a result, the CRPS approach is more computationally efficient, while still producing high-quality outputs statistically indistinguishable from the training data across various scales. As an aside, diffusion models also typically implement adaptive layer normalisation, but to condition on the timesteps in the de-noising process.

Neural networks often exhibit a spectral bias \cite{specbias}, leading to inadequate capture of higher frequency modes, which have a reduced contribution to overall energy compared to larger scales. In modelling physical data, it is crucial to preserve not only large-scale properties but also energy across all scales. To address this, we employ a spectral loss that utilises the discrete Fourier transform, $\mathcal{F}$, to transform $U_t$ and $\tilde{U}_t$. We then compute the modulus, yielding $|\mathcal{F}[U_{t}]| \in \mathbb{R} $ and $|\mathcal{F}[\tilde{U}_{t}]| \in \mathbb{R} $, before calculating the MAE between these two quantities. Through experimentation, we determined that setting $\lambda=1$ effectively weights these terms to achieve convergence during training.

Unlike VAEs, which incorporate a sampling layer through reparametrisation and typically enforce latent regularisation to shape the latent distribution $\mathcal{N}(\mu, \sigma)$ into a standard normal distribution $\mathcal{N}(0, 1)$, our approach does not impose such regularisation. Instead, the CRPS encourages ensemble diversity. It is important to note that the CRPS objective solely targets the marginal distributions, while generating spatio-temporally coherent samples requires matching the joint distribution of the data. Here the spectral bias term additionally plays a role in enhancing the model's ability to produce consistent and realistic spatio-temporal patterns 

While several approaches have explored using rollout training steps \cite{graphcast, FourCastNet} to achieve stable autoregressive rollouts, this adds additional computational overhead, and due to trajectory divergence in stochastic models making loss-based constraints less meaningful. Instead, our approach employs the composite loss described above, which facilitates stable autoregressive rollout training using only forecasts of \(\tilde{u}_{t+1}\), as demonstrated in Figure \ref{fig:KS_long}.

\subsection{Pre-training and Fine-tuning}
\label{chap:training}

The pre-trained weights serve as the initialisation for fine-tuning, where the model learns to generalise across dynamics associated with varying values of the PDE parameters. During fine-tuning, the initial optimisation learning rate is reduced to 1/50th of the pre-training learning rate. This ensures that the model does not deviate significantly from the solution obtained during pre-training. During pre-training, the scale and shift parameters are decoupled from the conditioning variables being fixed at $\gamma=1$ and $\delta=0$, which reduces computational complexity as there is no gradient flow between the objective function and $L $ or $\beta $. During fine-tuning, these parameters are no longer fixed but are instead dynamically adapted based on the conditioning values. This enables the model to incorporate the influence of varying PDE parameters, improving its ability to generalise across different dynamical regimes while preserving the representations learned during pre-training.

\section{Results}
\label{sec:results}

\subsection{Kuramoto-Sivashinsky Equation}

The Kuramoto-Sivashinsky (KS) equation is a canonical model for studying spatiotemporal chaos in dissipative systems \cite{unstableKS}  is given by:
\begin{align}
\partial_t u + u \partial_{x} u + \partial_{x}^2 u + \partial_{x}^4 u = 0,
\label{eq:ks_equation}
\end{align}
where $u(x,t)$ is a scalar field within a periodic domain $[0, L]$, where $L$ is the only parameter in the problem. $u \partial_x u$ describes nonlinear advection, while the second-order spatial derivative $\partial_{x}^2 u$ represents destabilising diffusion. The fourth-order spatial derivative $\partial_{x}^4 u$ introduces stabilising hyperviscosity, counteracting the destabilising diffusion and non-linear advection, thus leading to a balance that drives chaotic dynamics at large enough $L$.

Direct numerical simulations of the Kuramoto-Sivashinsky equation are conducted using a pseudo-spectral method with a fourth-order exponential time differencing Runge-Kutta solver using the FourierFlows package \cite{FourierFlows} in Julia \cite{Julia}, capable of running on the GPU. Simulations are initialised with a random field from a normal distribution, $u(x, 0) = \mathcal{N}(0, 0.01)$. After a warm-up phase, which is excluded from the training data, the system develops chaotic dynamics. The simulations are solved with a time step of \(\delta t=2.5\times10^{-2}\) and are temporally subsampled to $\Delta t = 1$ for training the emulator.

\begin{figure}[t]
\begin{center}
\includegraphics[width=\textwidth]{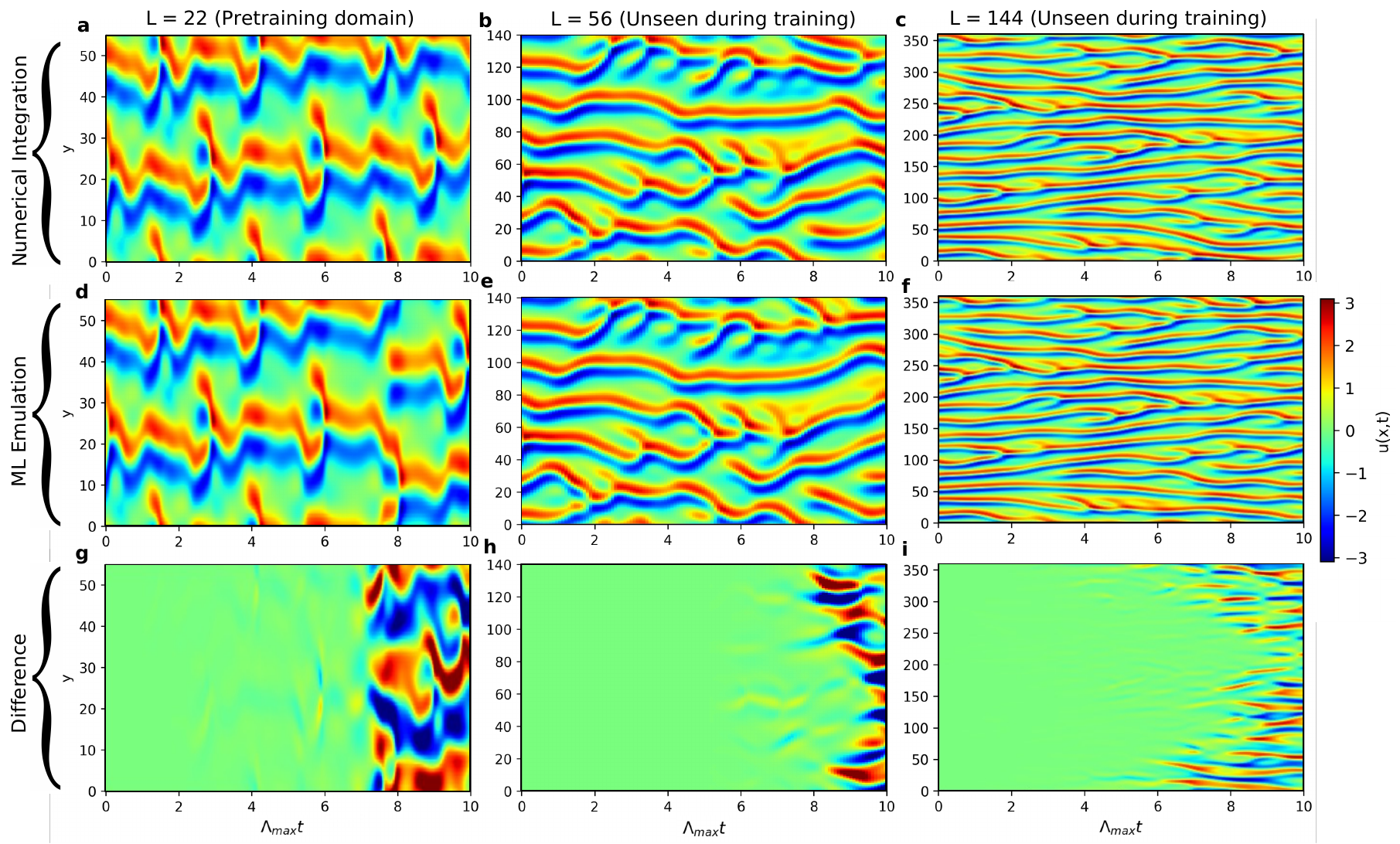}
\end{center}
\caption{Space-time plots for the Kuramoto-Sivashinsky (KS) equation, generated via numerical integration and ML emulation, from identical initial conditions. The bottom panel in each column displays the error between the upper two panels. We display results for $L=22$, which is used for pre-training, and $L=56, 144$, which are not shown to the network during training. Fine-tuning takes place for $L=\{36, 48, 64, 98, 128, 200\}$. The emulator successfully adapts to varying domain size $L$ and consequently input fields of varying sizes.}
\label{fig:KS}
\end{figure}

Previous work on emulating the KS equation has been limited to fixed domain sizes \cite{Linot, KS1, KS2} or requires \textit{ab initio} training for different $L$ values \cite{Pathek}. In our study, we employ a deep learning emulator to capture the dynamics of the KS equation across a range of domain sizes, $L$. As $L$ increases, the number of unstable modes in the system increases, leading to increased complexity and stronger chaotic behaviour. Consequently, the emulator must learn to generalise across different regimes of spatiotemporal chaos. As $L$ increases, computational costs also rise since $N_x$, the number of points in $x$, must be increased to resolve all active wavenumbers. While one could choose a sufficiently large $N_x$ to accommodate all values of $L$, this would lead to unnecessary computation for smaller values of $L$. Instead, by scaling $N_x$ with $L$, we ensure that only the minimum required number of grid points is used to resolve all relevant wavenumbers. However, this presents a challenge for machine learning approaches, which typically rely on fixed-size input-output pairs across all examples.

We develop an emulator for partial differential equations (PDEs) of the form $\partial_t U = F(U; L)$, where $F$ is a differential operator, and $L$ represents parameter values that influence the dynamics. We select a neural network to parametrise $F$, with parameters $\theta$, that is flexible to $x \in \mathbb{R}^{N_x}$ of any size. $F_{\theta}$ is designed to be both flexible enough to accommodate input fields of varying sizes and conditioned on the values of $L$. As detailed in Section 3, we base $F$ on local attention and adaptive layer normalisation to handle these requirements.

For the Kuramoto-Sivashinsky (KS) equation, the model is trained deterministically, following the standard practice in operator learning \cite{FNO}. Consequently, no sampling operation is implemented during training. Although training with the Continuous Ranked Probability Score (CRPS) results in similar error levels to deterministic training for the KS equation, the deterministic approach avoids the need for ensemble generation and instead uses the Mean Squared Error (MSE) as the loss function.

We pretrain the neural network on a weakly chaotic regime of the KS equation, domain size $L=22, x \in \mathbb{R}^{56}$. This allows for the bulk computation, using 5,000 snapshots, which occurs on the smallest fields. Fine-tuning takes place for $L=\{36, 48, 64, 98, 128, 200\}, u \in \mathbb{R}^{\{90, 120, 160, 320, 500\}}$ respectively, with 500 snapshots for each value. Here we see one of the advantages of implementing local attention as the network is adaptive to varying domain sizes.

We pretrain the neural network on a weakly chaotic regime of the KS equation, domain size $L=22, x \in \mathbb{R}^{56}$. This allows for the bulk computation, using 5,000 snapshots, which occurs on the smallest fields. Finetuning takes place for $L=\{36, 48, 64, 98, 128, 200\}, u \in \mathbb{R}^{\{90, 120, 160, 320, 500\}}$ respectively, with 500 snapshots for each value. Here we see one of the advantages of implementing local attention as the network is adaptive to varying domain sizes.

\subsubsection{Model Hyperparameters}

Using a Bayesian hyperparameter sweep \cite{wandb}, we determine the best-performing model during pre-training. The metric for minimisation was the MSE. Through this sweep, we found that the following hyperparameters were optimal: batch size: 128; optimiser: Adam; initial pre-training transformer learning rate: $ 5 \times 10^{-4}$; initial fine-tuning transformer learning rate: $10^{-5}$, pre-training epochs: 1000, fine-tuning epochs: 500, training ensemble size: 1; $S$ (length of time history used to make next forecast): 2, $C$ (latent channels): 64, local attention window size: 9, number of transformer blocks: 8, number of transformer heads: 1.

\subsubsection{Emulation}

The performance of our machine learning (ML) emulator for the Kuramoto-Sivashinsky (KS) equation demonstrates strong accuracy and generalisation across domain sizes. With the NN trained on a range of values of $L$, we compare model-generated trajectories with numerical simulations for various parameter values. To reduce training costs, the neural network is only fine-tuned on a limited selection of values of $L$ outlined above. We found this was sufficient as the neural network is capable of reproducing the expected dynamics on unseen parameter values when interpolating between seen values of $L$.

Figure \ref{fig:KS} presents latitude-time plots generated via numerical integration and ML emulation for different domain sizes. For the domain size $L=22$, which was used during pre-training, our emulator effectively captures the dynamics, exhibiting minimal error for $\sim7$ Lyapunov times compared to the numerical integration results. More notably, for previously unseen domain sizes $L=56$ and $L=144$, the emulator retains its ability to accurately reproduce the expected dynamics, with accurate tracking over $\sim7$ Lyapunov times, despite not having been explicitly trained on these parameter configurations.

\begin{figure}[t]
\begin{center}
\includegraphics[width=\textwidth]{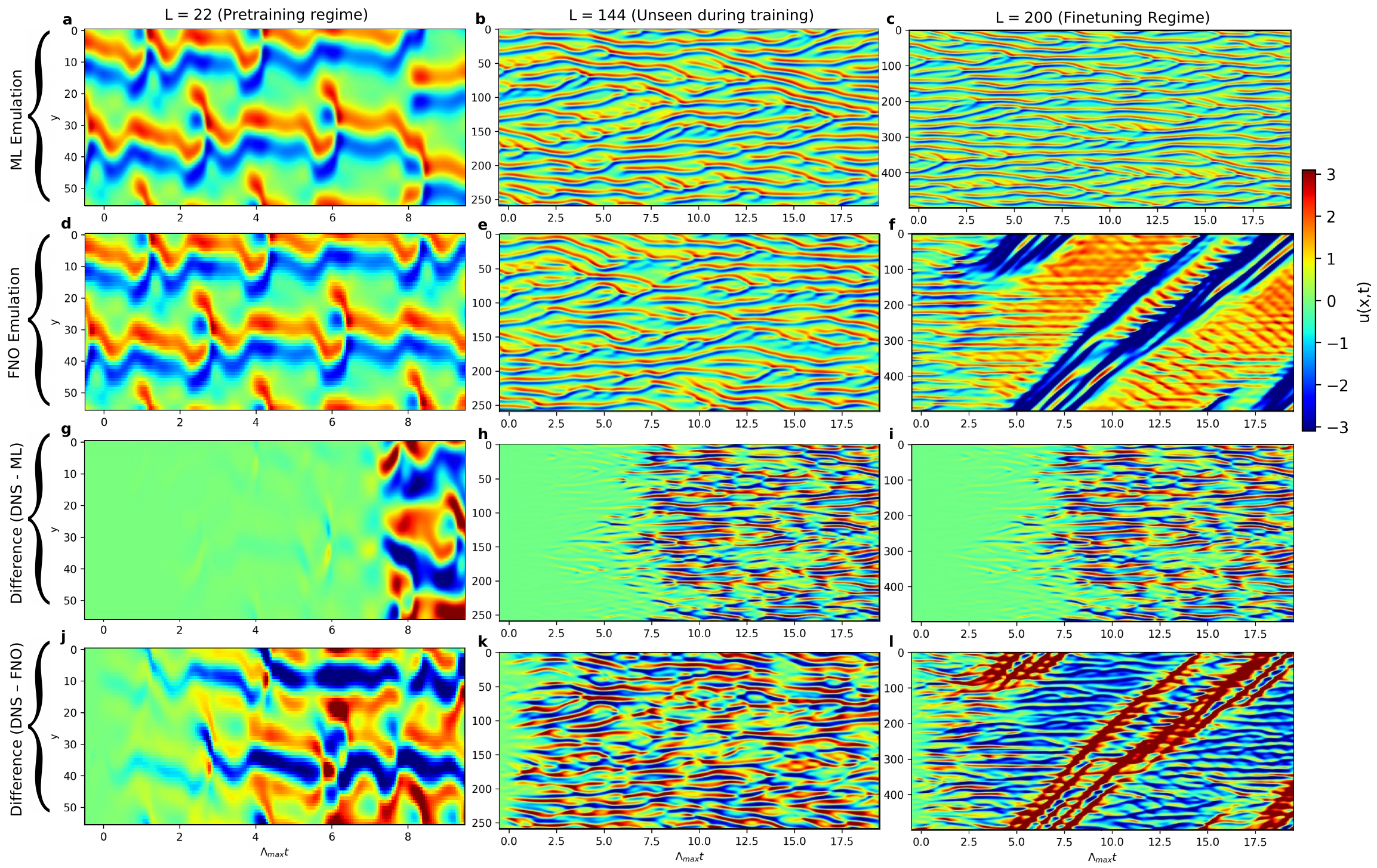}
\end{center}
\caption{Latitude-time plots for the Kuramoto-Sivashinsky (KS) equation, generated via the ML emulator and a FNO emulator, for comparison, from identical initial conditions. The bottom two panels in each column display the error between the upper two panels and the reference simulation. We display results for $L=22$, which is used for pre-training, and $L=144$, which is not shown to the network during training, and $L=200$, shown during fine-tuning, which takes place for $L=\{36, 48, 64, 98, 128, 200\}$. The FNO shows both poor short-term accuracy as well as an inability to model strongly chaotic behaviour, due to the threshold on the number of wavenumbers captured.}
\label{fig:KS_FNO}
\end{figure}

To compare with existing architectures, we examine the performance of the Fourier Neural Operator (FNO) across domains of varying sizes, as shown in Figure \ref{fig:KS_FNO}. While the FNO generates realistic forecasts for domain sizes $L=22$ and $L=144$ (Figures \ref{fig:KS_FNO}d and \ref{fig:KS_FNO}e), it struggles to produce a plausible trajectory when the domain size increases to $L=200$ (Figure \ref{fig:KS_FNO}f). This limitation arises because the FNO explicitly models only the largest $N$ wavenumbers, relying on a $1 \times 1$ convolution to recover higher wavenumbers. For instance, with $L=22$ and $N_x=56$ resolved grid points, the FNO models only the highest $N=28$ wavenumbers. However, when $L=200$, the system has approximately 31 unstable modes (estimated as $N_{\text{unstable}} \approx \frac{L}{2 \pi}$), exceeding the capacity of an FNO trained on $L=22$ to capture all dynamically unstable modes at the larger domain size. In contrast, the local attention mechanism explicitly models local dynamics by learning a translation-invariant stencil over the domain. This allows it to adapt flexibly to domains of arbitrary size while preserving the ability to resolve dynamics across scales.

\begin{figure}[t!]
\begin{center}
\includegraphics[width=\textwidth]{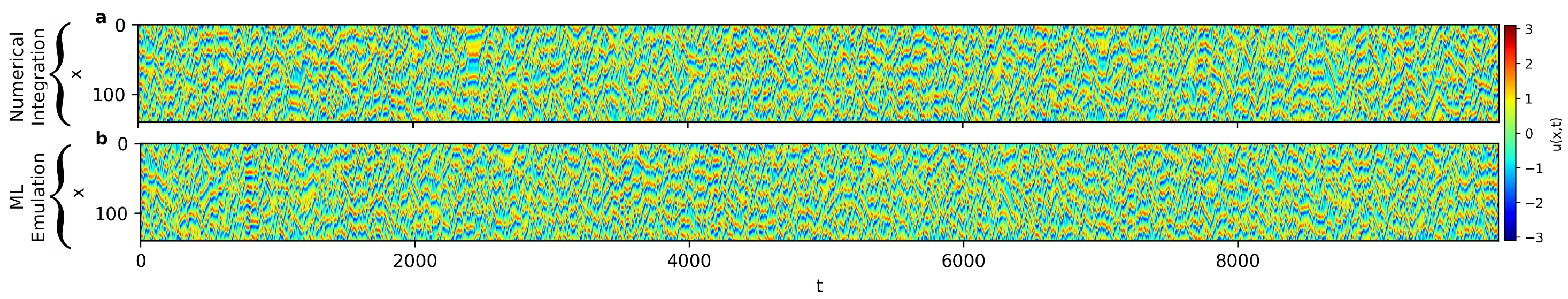}
\end{center}
\caption{Extended latitude-time plots for the Kuramoto-Sivashinsky (KS) equation for $L=56$, which is not shown to the network during training. Qualitatively, we see that the ML emulator is capable of remaining stable over very long autoregressive rollouts, here we display 10,000 steps. At each step the output from the network $\tilde{u}_{t+1}$ is used as the input to predict the next step.}
\label{fig:KS_long}
\end{figure}

To qualitatively assess the long-term stability of the ML emulator, we present latitude-time plots in Figure \ref{fig:KS_long}, illustrating the network's performance over an autoregressive rollout of 10,000 time steps. The emulator remains stable and consistent, reinforcing its ability to predict future states accurately. This suggests that the model’s forecasts are sufficiently accurate to avoid divergence from the system’s true distribution. In contrast, deviations from the true system can lead to out-of-distribution forecasts, often resulting in unrealistic trajectories. The comparison between neural network emulations and reference trajectories not only confirms the model's adeptness in capturing the system's dynamics but also highlights its robustness across a range of variable parameter values.

\begin{figure}[t]
\begin{center}
\includegraphics[width=\textwidth]{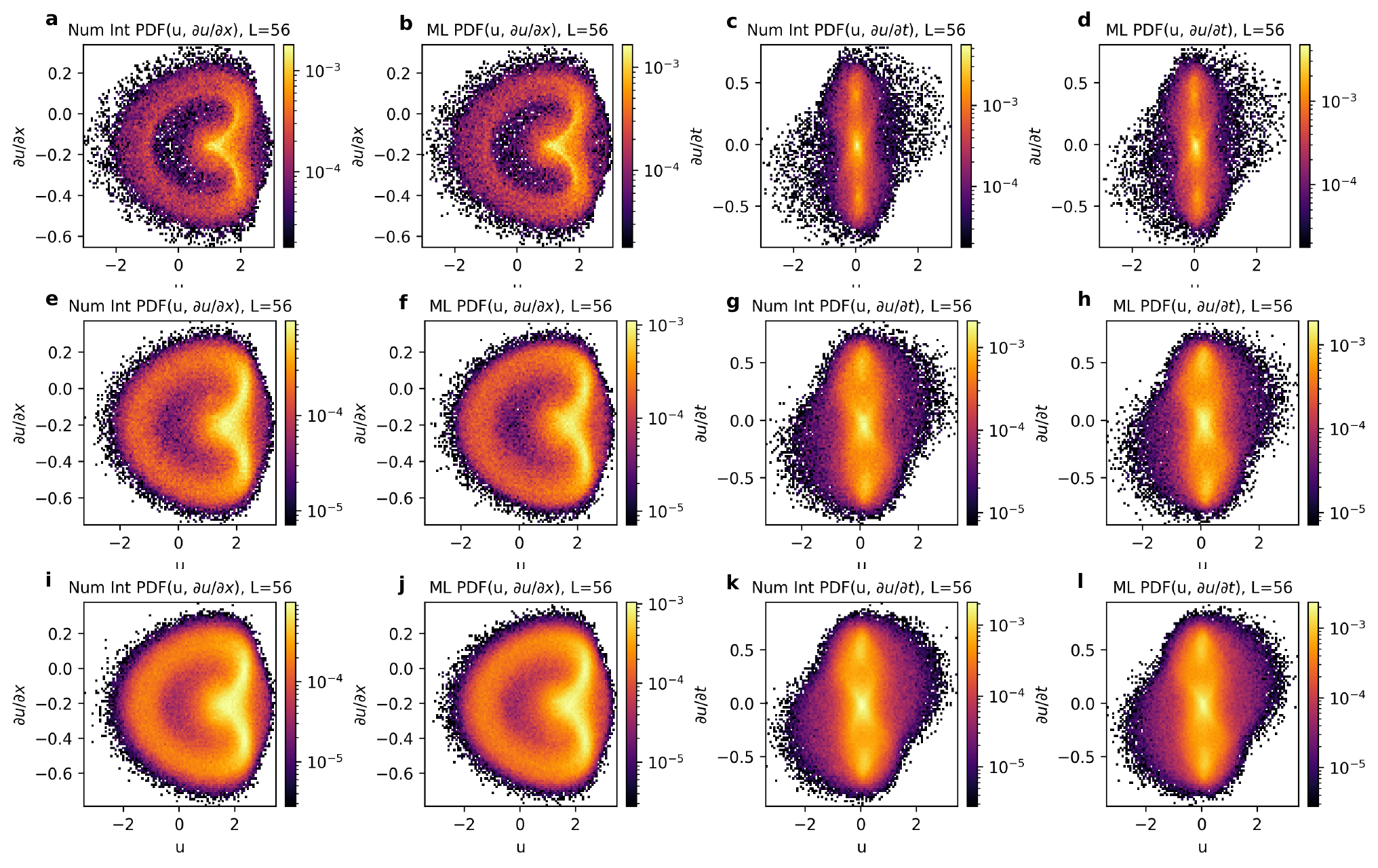}
\end{center}
\caption{PDFs displaying joint probabilities of $u$, $\partial_x u$ and $\partial_t u$ for the Kuramoto-Sivashinsky (KS) equation obtained over 1000 integration steps to observe statistical properties of each regime. \textbf{a}, \textbf{e}, \textbf{i}, for visualisation of the 3D PDF we average over $\partial_t u$ to obtain the density of values in the $u$-$\partial_x u$ space for different values of $L=22$, used for pre-training and with $L=\{56, 144\}$, unseen during training. \textbf{b}, \textbf{f}, \textbf{j} show the same $u$-$\partial_x u$ space from an evolution generated using the neural network. \textbf{c}, \textbf{g}, \textbf{k} show the $u$-$\partial_t u$ space by averaging over $\partial_x u$ for the numerical integration values and \textbf{d}, \textbf{h}, \textbf{l} show the $u$-$\partial_t u$ space for evolution generated using the neural network. Here we can see that the emulator shows very good agreement with the numerical integration, across the parameter range.}
\label{fig:KS_pdf}
\end{figure}

For a quantitative assessment of longer-term evolutions, we examine statistical properties of the system. We generate probability density functions (PDFs) for individual $u$ values and their derivatives, $\partial_x u$ and $\partial_t u$. These PDFs serve to measure the learned spatial and temporal correlations. We conduct this analysis over a time integration period of $1000$ time units, producing two PDFs: $p(U)$ representing the PDF from numerical integration, $U$, and $q(\tilde{u})$ representing the PDF of forecasted values derived from the ML emulator, $\tilde{u}$. In Figures \ref{fig:KS_pdf} we plot the joint PDFs of $u$, $\partial_x u$ and $\partial_t u$. For visualisation of the 3D PDFs, we sum over $\partial_tu$ to obtain the density of values in the $u$-$\partial_x u$ space, shown in Figure \ref{fig:KS_pdf}a,e,i obtained via numerical integration and Figure \ref{fig:KS_pdf}b,f,j obtained from the neural network. Similarly, we show the density of values in the $u$-$\partial_tu$ space, by summation over $\partial_x u$, for numerical integration in Figure \ref{fig:KS_pdf}c,g,k and the neural network in Figure \ref{fig:KS_pdf}d,h,l. The visualisations indicate that the ML emulator reproduces the joint distributions observed in the numerical integration, validating its capability to model the statistical behaviour of the KS equation across the parameter space effectively.

\subsection{Beta-Plane Turbulence}
\label{chap:eds_beta}

We now consider the application of our deep learning emulator to the well-studied idealized Geophysical Fluid Dynamics (GFD) system—beta plane turbulence \cite{rhines_1975, zonal_jets}—spanning a range of system parameters. This system serves as a useful model for various atmospheric and oceanic flow aspects. Modelled as a single-layer flow on a doubly-periodic domain, turbulence is generated via stochastic forcing \(\xi(x, y, t)\) in the vorticity equation:
\begin{align}
\partial_t \zeta + u \cdot \nabla \zeta + \beta \partial_x \psi = \xi - \mu \zeta + \nu_n \nabla^{2n} \zeta,
\label{eq:beta_plane}
\end{align}

Here, the velocity field is defined as \(u(x, y, t) = (-\partial_y \psi, \partial_x \psi)\) (following the common GFD sign convention), with \(\psi\) representing the stream function and \(\zeta = \partial_x v - \partial_y u\) denoting relative vorticity. Energy dissipation occurs through linear damping at a rate \(\mu\), while hyperviscosity, characterised by coefficient \(\nu\) and order \(n\), removes vorticity at small scales. The forcing term \(\xi(x, y, t)\) consists of white-in-time noise injected into an annulus in wavenumber space (see \cite{SLT} for details). The parameter \(\beta\), representing the latitudinal variation of the Coriolis parameter, is varied to examine the system's sensitivity to it.

Direct numerical simulations of equation (\ref{eq:beta_plane}) are conducted on a 2D doubly-periodic square domain \((x,y) \in [0, 2\pi]^2\) at a resolution of \(N_x = N_y = 256\) using a pseudo-spectral method with a fourth-order Runge-Kutta solver in Julia \cite{Julia} using the GeophysicalFlows package \cite{GFF} on GPU. A wavenumber-filter is applied at every time-step which removes enstrophy at high wavenumbers. The stochastic forcing \(\xi(x, y, t)\) is injected onto an annulus of wavevectors in Fourier space centred around a mean radial wavenumber \(k_f\) with thickness \(\delta k = 1\). In all experiments, we set \(\mu=4\times10^{-2}\), \(\nu=1\), and an energy injection rate of \(\varepsilon=10^{-4}\), with a principal forcing wavenumber of \(k_f=16\) and a time step of \(\delta t=4\times10^{-2}\). These parameters are chosen to ensure the formation of coherent zonal jets, exhibiting a wide range of temporal variability. Simulations are initialized from rest with \(\zeta = 0\), and parameters remain constant throughout the integration. The time required for the total kinetic energy to reach a statistically steady state is influenced by the damping rate \(\mu\), typically achieved by \(\mu t = 2 - 3\) (here, \(t = 50 - 75\) time units).

In turbulence modelling, a Reynolds decomposition is often employed to separate coherent structures from fluctuations, which here is applied as an eddy-mean decomposition in the zonal direction: \(u(x,y,t) = \overline{u}(y,t) + u'(x,y,t)\) (where \(\overline{u}(y, t)=\frac{1}{L}\int^{L}_{0}u \ dx\) and \(L=2\pi\) is the domain size in \(x\)). This leads to equations for the zonally-averaged velocity field \(U(y, t)=\overline{u}(y, t)\) and the associated eddy fluctuation fields \(\zeta^{'}(x, y, t)\):
\begin{align}
\partial_t U&= \left(-\mu + \nu_n \partial_y^{2 n}\right) U + \overline{\zeta^{'} v^{'}}, 
\label{eq:dU_dt}\\
\partial_t \zeta^{'}&= \left(-\mu + \nu_n \nabla^{2 n} -U \partial_x\right)  \zeta^{'} + \left(\partial_{y y} U-\beta\right) v^{'} + \xi + {EENL}, 
\label{eq:dzeta_dash_dt}
\end{align}
where \({EENL}:= \partial_y(\overline{\zeta^{'} v^{'}})-\partial_y(\zeta^{'} v^{'}) -\partial_x(\zeta^{'} u^{'})\) denotes the eddy-eddy non-linear interaction terms. The evolution of \(U\) depends on the non-linear interaction between the eddies and the mean flow, encapsulated by the Reynolds stress term \(\overline{\zeta^{'} v^{'}}\). The forcing noise \(\xi\) is applied solely to the eddy fields, influencing the mean flow indirectly through non-linear processes. Similarly, the influence of \(\beta\) also manifests indirectly via the Reynolds stress term.

By introducing this decomposition, we shift away from operator learning, which seeks to model the time evolution of fully-resolved fields, such as \(\zeta(x,y, t)\). Instead, by modelling the under-resolved system \(\partial_t U(y, t)\) with our emulator, the influence of unresolved variables on \(U\) must implicitly be learned from data. This approach aligns with broader GFD modelling efforts, where data from observations and general circulation models (GCMs) often comes in the form of fields with coarse resolution, when compared to the scales in which processes take place.

The probabilistic variant of the deep learning emulator can be trained for PDEs of the form \(\partial_t U = F(U; \beta) + G(U, u'; \beta)\), where \(G\) parameterises the interaction between resolved variables \(U\) and unresolved variables \(u'\). The latter is governed by \(\partial_t u' = H(U, u'; \beta) + \xi\), where \(F\), \(G\), and \(H\) are operators and \(\xi\) is a noise term. This formulation captures the dynamics of beta plane turbulence, as represented in equations (\ref{eq:dU_dt}) and (\ref{eq:dzeta_dash_dt}), as well as many other under-resolved systems, such as fluid flows in the atmosphere and oceans.

\begin{figure}[t]
\begin{center}
\includegraphics[width=\textwidth]{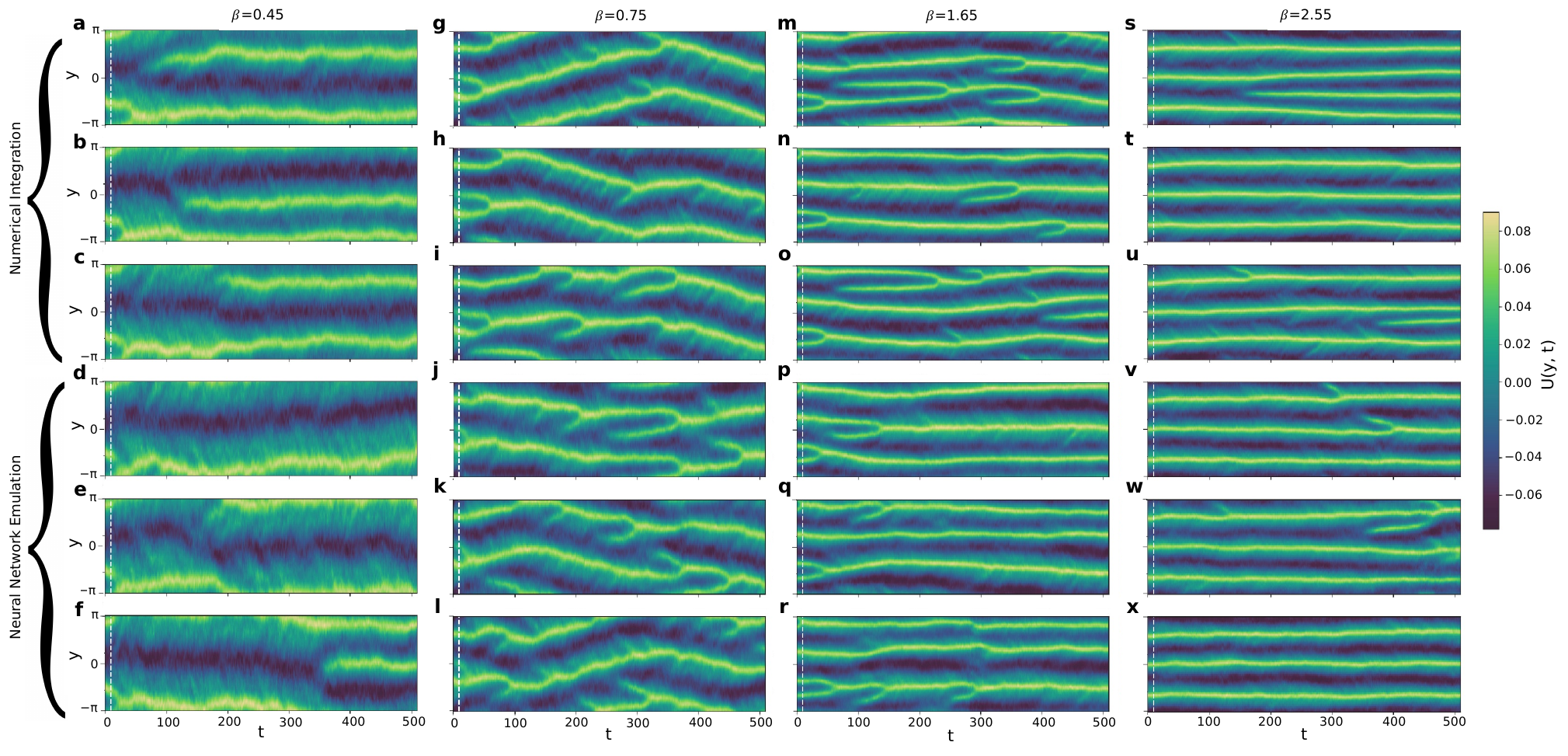}
\end{center}
\caption{Latitude-time plots of the zonally-averaged zonal flow $U(y,t)$ displaying an ensemble of a numerical integration and neural network emulations across previously unseen parameter values. The neural network is initially pre-trained for $\beta=0.9$ using 200,000 snapshots, followed by fine-tuning for $\beta=\{0.3, 0.6, 1.2, 1.5, 1.8, 2.1, 2.4, 2.7\}$ with 5,000 snapshots for each value. The neural network produces an ensemble of trajectories originating from identical initial conditions, showcased here over a duration of 500 time units, for the previously unseen parameter values of $\beta=\{0.45, 0.75, 1.65, 2.55\}$. The neural network produces qualitatively plausible trajectories, adeptly generalising to the different regimes, capable of learning the system dependence on $\beta$.}
\label{fig:lat_time}
\end{figure}

As with most PDEs, varying the governing equation parameters gives rise to diverse phenomena. In this system, changing the parameter \(\beta\) results in a range of dynamics, illustrated in the latitude-time plots in Figure \ref{fig:lat_time}. Not only does the number of jets (maxima in \(U\) represented in yellow) vary, but the dynamic tendencies, including the frequency of new jet nucleation, coalescence of jets, and latitudinal jet translation, also change.

Since the neural network only accesses the resolved variables \(U\), we provide two previous time steps, \(U_{t-1}\) and \(U_t\) (S=2 in Figure \ref{fig:ml_arch}). To learn a probabilistic mapping, we implement a probabilistic loss function, which requires only a single forward pass through the model. This contrasts with diffusion methods \cite{diffusion}, which typically require 20 to 50 iterations through the model to generate a forecast for \(U_{t+1}\).

In the case of the beta-plane turbulence, all values of $\beta$ were simulated with $N_x = N_y = 256$. The neural network is initially pre-trained with a dataset of 10,000 snapshots where $\beta=0.9$, followed by fine-tuning with $\beta=\{0.3, 0.6, 1.2, 1.5, 1.8, 2.1, 2.4, 2.7\}$ with 1,000 snapshots for each value. This range was selected because it is characterised by well-defined and strongly variable jets.

\subsubsection{Model Hyperparameters}

We again use a Bayesian hyperparameter sweep \cite{wandb} to determine the best-performing model during pre-training. The metric for minimisation was the composite loss function outlined in \ref{eq:loss}. Through this sweep we found that the following hyperparameters were optimal: batch size: 128; optimiser: Adam; initial pre-training transformer learning rate: $5 \times 10^{-4}$; initial fine-tuning transformer learning rate: $10^{-5}$, pre-training epochs: 500, fine-tuning epochs: 250, CRPS training ensemble size: 2; $S$ (length of time history used to make next forecast): 2, $C$ (latent channels): 64, local attention window size: 17, number of transformer blocks: 16, number of transformer heads: 1.

\subsubsection{Emulation}

It is not possible to accurately predict a single trajectory due to the stochastic nature of the system, but it is possible to generate an ensemble of equally-likely trajectories. With the neural network trained over a range of values of $\beta$, we compare an ensemble of ML-generated trajectories with an ensemble of numerical simulations for various parameter values. Figure \ref{fig:lat_time} displays the zonally-averaged zonal flow, for four values of $\beta$ unseen during training, with each regime exhibiting distinct dynamics. The neural network emulations closely align with reference numerical solutions, with accurate representation of jet nucleation and coalescence frequencies, as well as the latitudinal translation rate. Importantly we observe that the neural network does not become physically inconsistent after a number of autoregressive steps.

Comparisons between neural network emulations and reference trajectories confirm the model's skill in capturing system dynamics and demonstrate its robustness across a range of parameter values. To quantitatively assess the neural network’s performance, we examine its ability to replicate key statistical properties of the stochastically driven jet system. These properties include spatial and temporal correlations, as observed in the KS equation, along with spectral characteristics—a crucial metric for accurately modelling flows in GFD \cite{gencast}.

\begin{figure}[t!]
\begin{center}
\includegraphics[width=\textwidth]{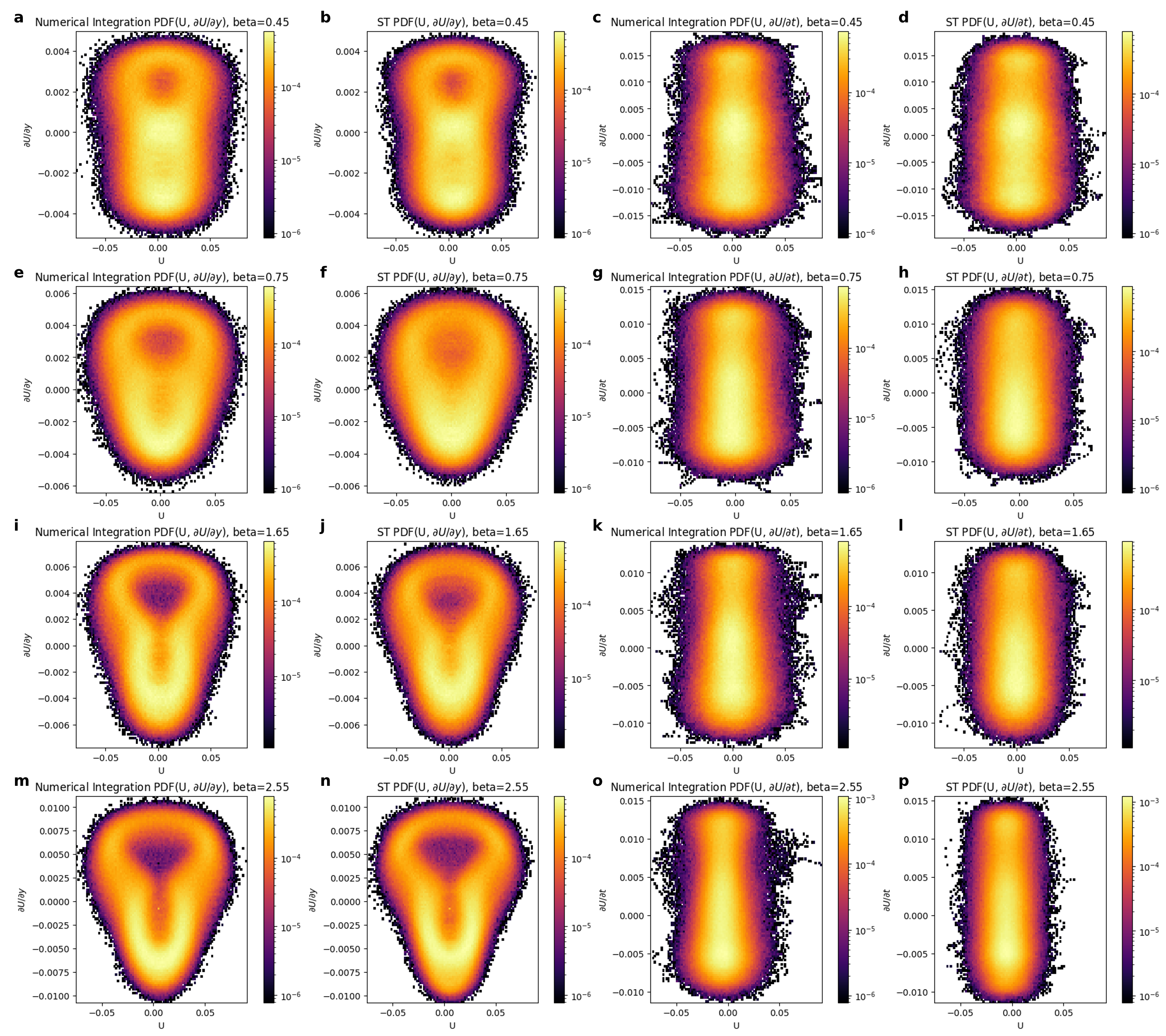}
\end{center}
\caption{PDFs displaying joint probabilities of $U$, $\partial_y U$ and $\partial_t U$ for the beta plane system obtained over 5000 integration steps. The plots take the same form as in Figure \ref{fig:KS_pdf}, for values of $\beta=\{0.45, 0.75, 1.65, 2.55\}$, all unseen during training. Here we can see that again the neural network shows good agreement with the numerical integration, covering the full space of dynamics, across the parameter range.}
\label{fig:val}
\end{figure}

In Figure \ref{fig:val}, we observe strong agreement between the outputs of the neural network and the numerical integration across different parameter values, effectively capturing the full range of dynamics. However, as $\beta$ increases, discrepancies emerge in the joint PDFs of $U-\partial_t U$ generated by the neural network and numerical integration. These discrepancies may stem from a reduced frequency of spontaneous events as zonostrophy increases (i.e., the tendency for zonal flows to dominate over eddies), which could lead to data imbalance. While this phenomenon warrants further investigation, the overall agreement between the numerical integration and neural network emulations remains excellent.

\begin{figure}[t!]
\begin{center}
\includegraphics[width=\textwidth]{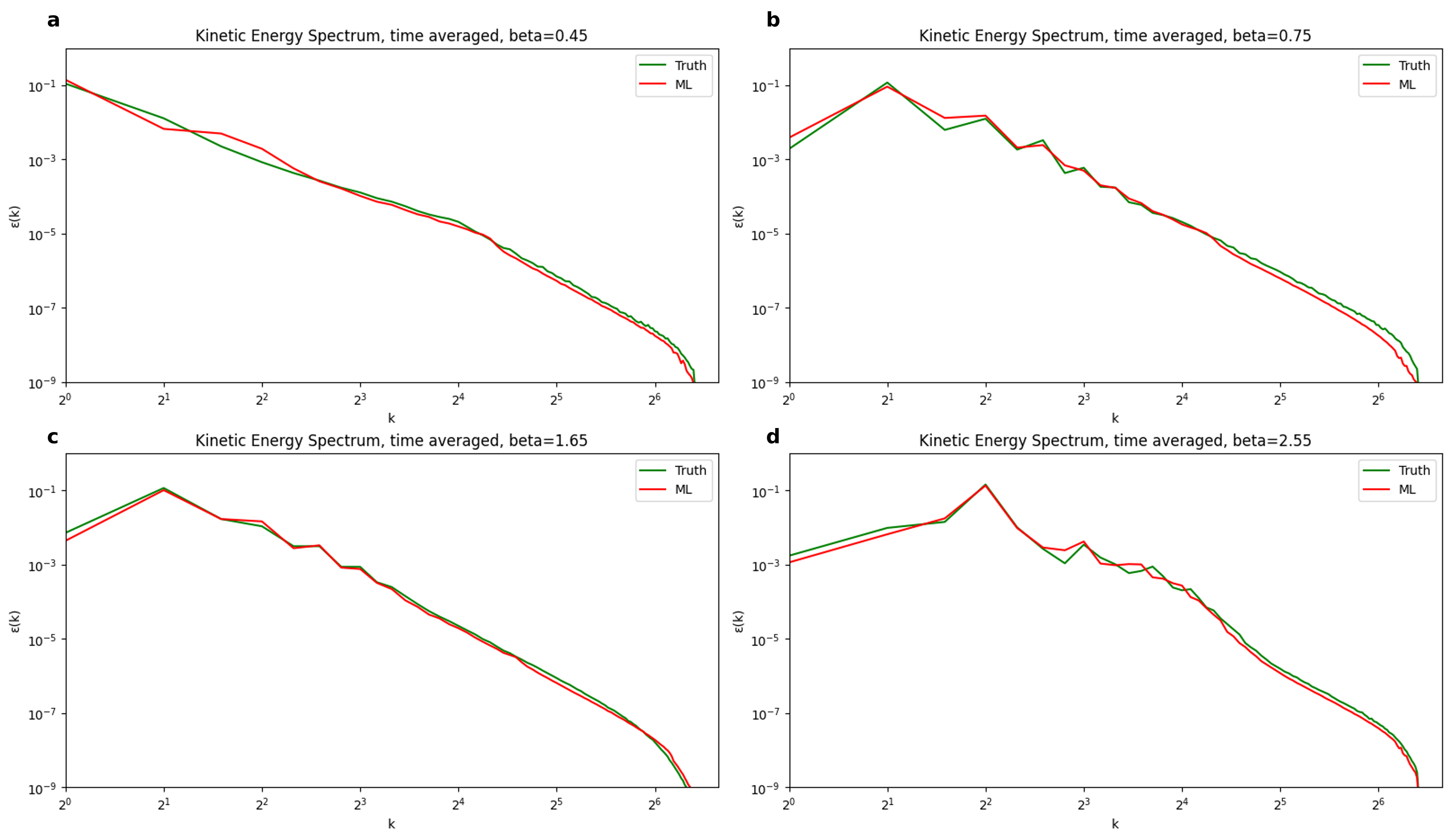}
\end{center}
\caption{Time averaged zonal power spectral density displayed for both the numerical integration (green) and the neural network (red) for the beta plane system, showing the energy content of each wavenumber in the zonally-averaged zonal direction, $U$, for values of $\beta=\{0.45, 0.75, 1.65, 2.55\}$, averaged over 5,000 time units. Again the neural network shows good agreement with the numerical integration for values of $\beta$ unseen during training.}
\label{fig:spectra}
\end{figure}

The proportion of energy in each wavenumber, or the energy spectra, is an important physical property in fluid dynamics. For this we assess time-averaged power spectral density plotted for both the numerical integrations (in green) and the neural network (in red) over the same time integration period of $5000$ time units as used above, plotted in Figure \ref{fig:spectra}. We observe a high level of agreement across all scales, paying particular attention on the higher wavenumbers where data-driven approaches often fail. This is all the more important in the case of 2D turbulence due to an inverse energy cascade. This means that correct representation of the small scales is vital due to their upstream influence on the large scale dynamics.

These findings, supported by quantitative evaluation metrics, confirm the neural network’s capacity to accurately emulate system dynamics across a wide range of parameter values, including those unseen during training. Our results show that the ML emulator not only adapts to varying domain sizes and parameter values but also maintains high accuracy and consistency in long-term predictions. This robustness, combined with its computational efficiency, positions the ML approach as a powerful tool for investigating complex dynamical phenomena in beta plane systems and augurs well for advancing fluid dynamics research and its applications.

\subsubsection{Uncertainty Quantification: Spontaneous Transition Events}

\begin{figure}[t!]
\begin{center}
\includegraphics[width=\textwidth]{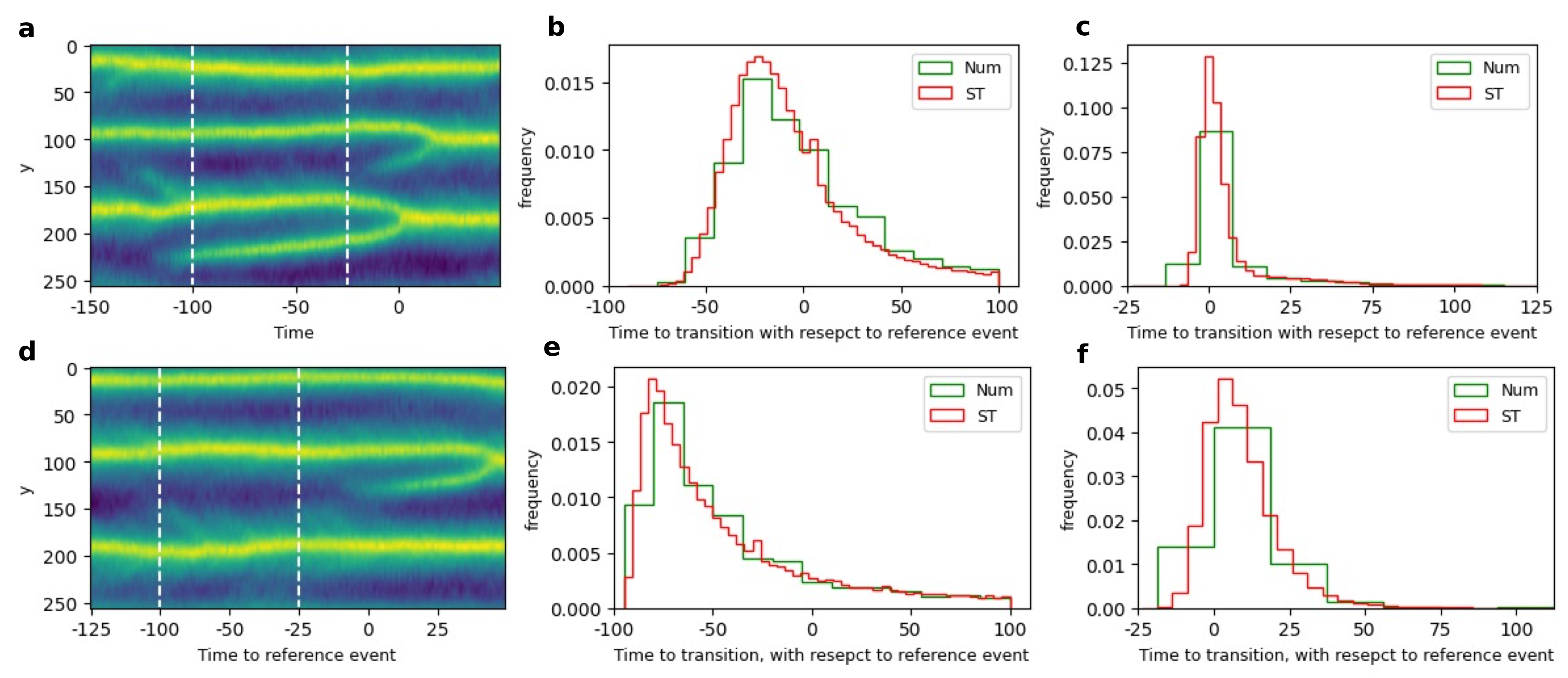}
\end{center}
\caption{\textbf{a}, Reference coalescence event, obtained via numerical integration of the beta plane system with $\beta=1.35$, a parameter value not seen by the neural network during training. The two dotted lines indicate initial conditions used to initialise ensembles via numerical integration and the neural network. \textbf{b}, Probability density functions (PDFs) showing the time to a coalescence event, with respect to the reference event in shown \textbf{a}, with the initial conditions 100 time units before the reference coalescence event. The green curve shows the PDF from numerical integration, comprising an ensemble of size 1000. The red curve shows the PDF from the neural network, comprising an ensemble of 100,000. \textbf{c}, Shows the same information as \textbf{b}, but for initial conditions 25 time units before the reference coalescence event. \textbf{d}-\textbf{f}, Show the same information as \textbf{a}-\textbf{c}, but for a reference nucleation event. We observe agreement between the green and red curves indicating that the ML emulator is successfully replicating the expected distributions.}
\label{fig:nuc_coal}
\end{figure}

In chaotic or stochastically driven systems predicting transition events, such as nucleation and coalescence events, are of interest; however, large ensembles are required for accurate estimation of transition probabilities. As outlined in \cite{SLT}, the neural network provides a five-order-of-magnitude speed-up over traditional numerical methods, facilitating the generation of such ensembles.

Figure \ref{fig:nuc_coal}a displays a reference coalescence event, while Figure \ref{fig:nuc_coal}d displays a reference nucleation event, both generated via numerical integration with $\beta=1.35$, a parameter value not seen during training. Each plot shows dashed lines prior to each event to display the initial conditions used to generate ensembles and measure the time interval between each initial condition and event of interest. In Figures \ref{fig:nuc_coal}b,c,e,f the neural network generates PDFs that closely match those obtained via numerical integration. This demonstrates the neural network's capability of effectively capturing the dynamics of the physical system while generalising to unseen regimes. This allows for computationally efficient exploration of physical systems where previously prohibitively expensive.

Sampling the network in this manner provides an estimate of aleatoric uncertainty, with the model naturally demonstrating reduced uncertainty as the forecast approaches the event occurrence. Moreover, it exhibits lower uncertainty when predicting coalescence events compared to nucleation events, consistent with observations from numerical studies in \cite{Laura_cope}.

\section{Conclusion}
\label{sec:conclusion}

Our neural network emulator effectively captures the dynamics of the Kuramoto-Sivashinsky (KS) equation and the stochastically-driven beta-plane system, demonstrating its ability to generalise across unseen parameter values. We demonstrate its ability to accurately reproduce key statistical properties, including spatial and temporal correlations as well as spectral characteristics. The efficiency of the emulator enables rapid exploration of parameter spaces and enables the investigation of complex phenomena, such as spontaneous transitions.

These results demonstrated successful interpolation within finely tuned parameter ranges. Further exploration of the emulator’s generalisation capabilities — in particular, zero-shot predictions of dynamics with extrapolated parameter values — has been carried out in related work \cite{shokar_extreme}. That study assessed its effectiveness in forecasting unseen events and its ability to characterise dynamical statistics from limited training data.

Exploring the parameter space is crucial in nearly all dynamical systems, and this framework offers a computationally efficient approach. By integrating local attention with adaptive layer normalisation, the emulator adapts flexibly to different domain sizes, enabling efficient pre-training on smaller domains while generalising effectively to larger ones. In this work, the architecture was also expanded to handle 2D spatial interactions (with vertical layers treated in the channel dimension), broadening its applicability to diverse fluid systems and high-dimensional dynamics in both deterministic and stochastic contexts.

For future work, the emulator could be extended to explore higher-dimensional parameter spaces by conditioning on multiple PDE parameters. Such an approach would enable a more comprehensive characterisation of system behaviour across the parameter landscape, supporting the efficient identification of critical parameters and bifurcations that govern jet-structure transitions and spontaneous-event frequencies. This direction holds promise for accelerating data-driven emulation of complex systems, particularly where generating large datasets for high-resolution domains is prohibitively expensive.

\paragraph{Acknowledgments}
We are grateful to the organisers, reviewers, and participants of the 2024 ICLR Workshop on AI for Differential Equations in Science. We also thank Miles Cranmer and Pedram Hassanzadeh for their thoughtful reviews of this work and their comments on the broader thesis. Their insights and feedback have been invaluable in strengthening this study.

\paragraph{Funding Statement}
I.S. acknowledges funding by the UK Engineering and Physical Sciences Research Council (grant number EP/S022961/1) as part of the UKRI Centre for Doctoral Training in Application of Artificial Intelligence to the Study of Environmental Risks.

\paragraph{Competing Interests}
No competing interests.

\paragraph{Data Availability Statement}
The code used in this work will be made publicly available upon publication at \url{https://github.com/Ira-Shokar/Stochastic-Transformer}

\paragraph{Ethical Standards}
The research meets all ethical guidelines, including adherence to the legal requirements of the study country.

\bibliographystyle{ieeetr}
\bibliography{references}

\newpage

\appendix

\section{Local Attention}

Local attention focuses on a spatially localized window around each point, akin to the receptive field in convolutional networks. It captures small-scale interactions, while global structures are handled across layers. By leveraging attention mechanisms instead of traditional convolutional operations \cite{cnn}, this approach dynamically computes spatial weights, ensuring adaptability to varying input scales. 

The input $u \in \mathbb{R}^D$, where $D$ represents the number of spatial grid points, is encoded into a latent space $z$ for transformer operations. This encoding is performed via a linear transformation with weights $W_{\text{E}} \in \mathbb{R}^{1 \times C}$, where $C$ denotes the number of channels. After processing through the transformer blocks, a decoder applies a linear transformation that maps the $C$ channels back to a single channel, yielding the prediction $\tilde{u}_{t+1} \in \mathbb{R}^D$. The overall architecture is schematically depicted in Fig.5.

\begin{figure}[t!] 
\centering
    \includegraphics[width=0.5\textwidth]{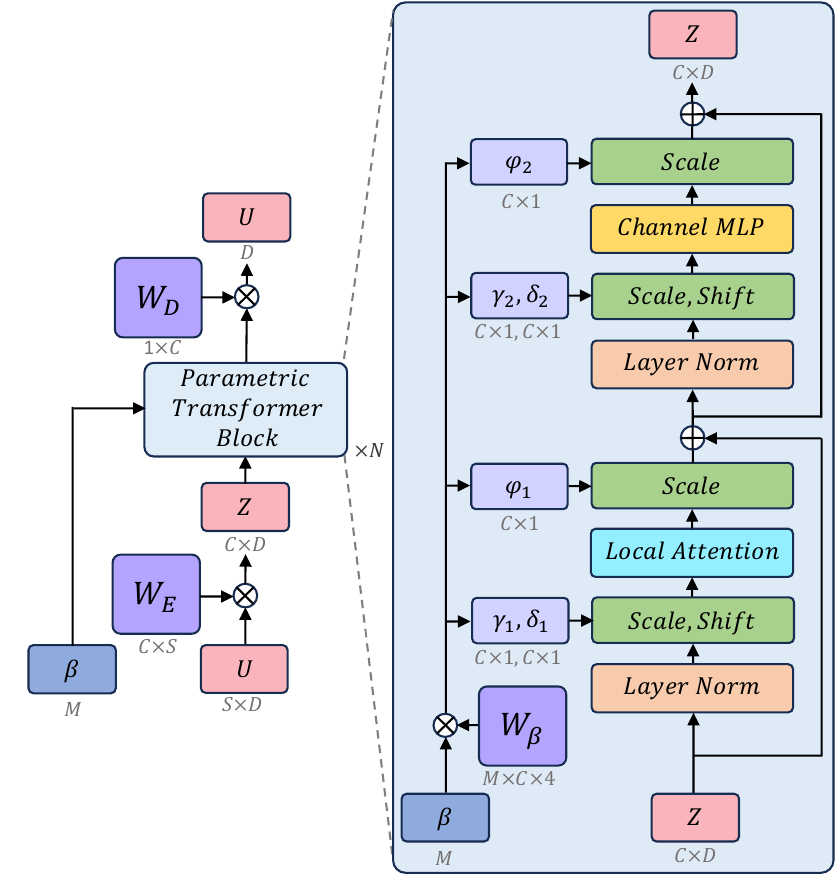}
        \caption{
       Schematic of the NN architecture. The network is structured around a transformer architecture conditioned on parameter $L$ and initialised by conditions $U$. Within each transformer block, adaptive layer normalisation conditions the transformer on $L$ by replacing scale and shift parameters. Each $W$ denotes learned weights for linear transformations, with arrows indicating the forward pass. In this study, the conditioning parameter $L$ is defined with a size of $M=1$ and the temporal history provided to the model $S=1$. However, the architecture is flexible to extension to larger dimensions within the parameter space.
        }
    \end{figure} 

The local attention layer computes correlations within a neighbourhood $\mathit{N}_K(x)$ of spatial extent $K$ around each position $x$ in $z \in \mathbb{R}^{C \times D}$. For $x$, attention is computed as $a_x = \sum_{b \in \mathit{N}_K(x)} \text{softmax}_b(\zeta_x^T k_b) v_b$, where $q$, $k$, and $v \in \mathbb{R}^{C \times D}$ are query, key, and value vectors from linear projections of $z$ \cite{Attention}. This mechanism extracts local contextual information, enhancing focus on relevant neighbourhood features. In our implementation, we also employ circular padding to ensure that the neighbourhoods wrap around at the boundaries of the input tensor, to retain translational equivariance. To encode relative positional information between elements, we introduce a learnable relative positional encoding, $\text{PE} \in \mathbb{R}^{K}$. This approach achieves a large receptive field with fewer layers, avoiding pooling or striding. It is especially beneficial for multi-scale dynamics. By using relative positions and circular padding in the unfold operation, the attention mechanism preserves translation equivariance, aligning with the symmetries of the dynamics.

\subsection{Parametric Transformer}

\begin{figure}[t!] 
\centering
    \includegraphics[width=\textwidth]{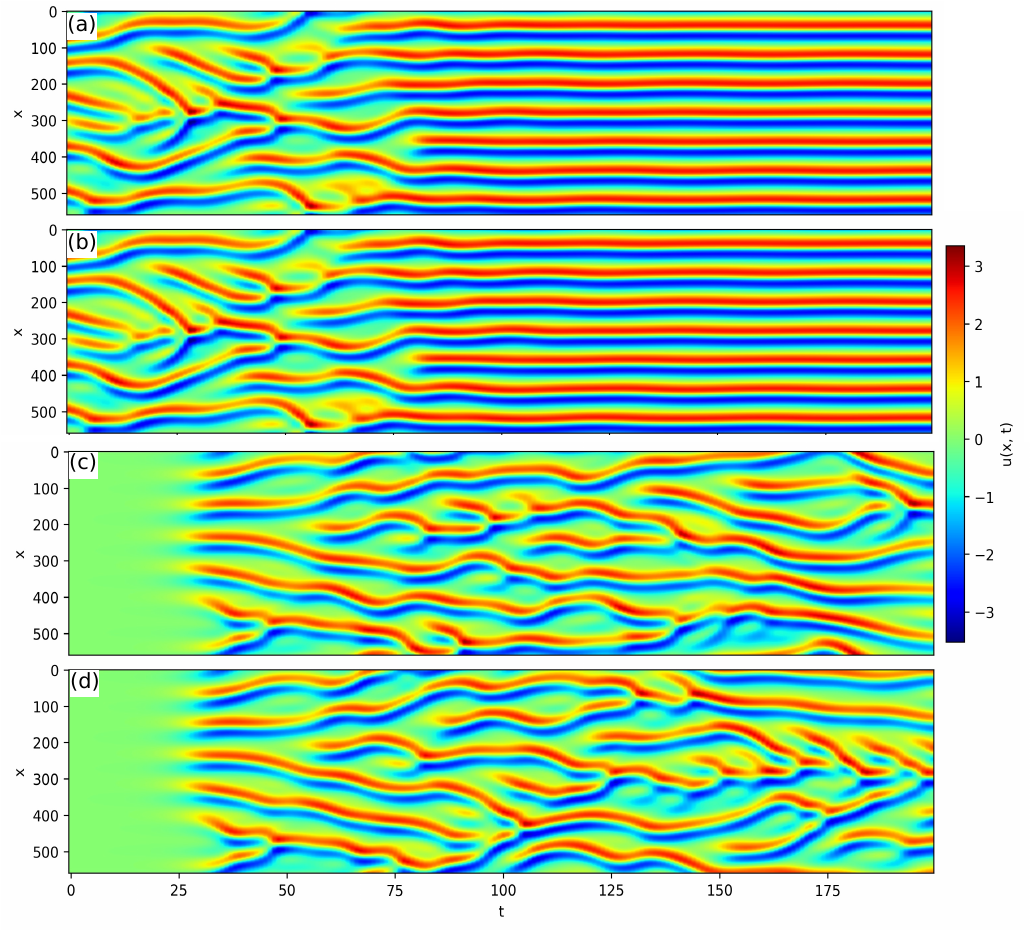}
        \caption{(a) Relaminarisation event from Fig.1 (a) captured by a single layer fully-connected NN trained and evaluated on $L=56$, where no laminar dynamics were present in the training data. (b) Same as (a) generated by FNO. (c) Initialisation of KS flow from initial conditions from Fig.1 (d) emulated by a single layer MLP trained and evaluated on $L=56$. (d) Same as (c) for the FNO.}
\end{figure} 

\begin{figure*}[t!] 
    \centering
    \includegraphics[width=\textwidth]{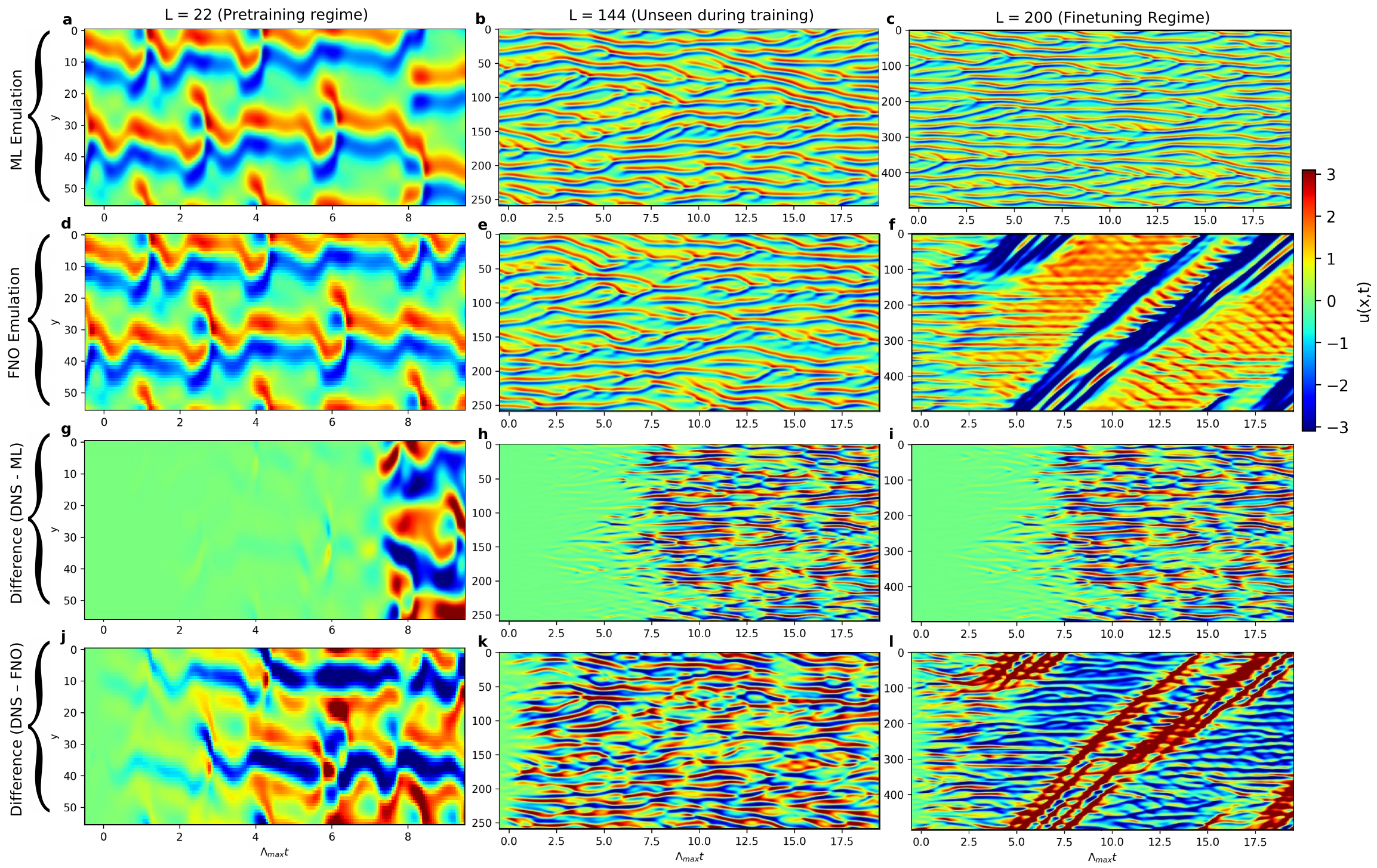}
    \caption{Generalisation capability of the Neural Network. All are pretrained with $L=22$ before finetuning on $L=\{22, 36, 48, 64, 98, 128, 200\}$. (a.-c.) Local Attention based NN. (d.-f.) FNO.}
\end{figure*}

We replace the affine parameters in layer normalisation with a learned function of the conditioning information \cite{FILM}. This approach enables leaner pre-training by decoupling scale and shift parameters from the conditioning variables, improving transformations during fine-tuning. We define scaling ($\gamma$) and shift ($\delta$) parameters to modulate activations in both the attention operation and MLP, totalling four parameters. These are obtained via a linear transformation $W_\beta \in \mathbb{R}^{M \times 4C}$ applied to the context vectors $\beta$, where $M=1$ represents their size. While this work focuses on $M=1$, the framework is easily extendable to higher-dimensional parameter spaces. Transformations within each transformer block are given by:

\begin{align}
    z &\rightarrow z + \varphi_1 \cdot \mathrm{LA}\left(\gamma_1 \cdot \mathrm{LN}(z) + \delta_1\right), \\
    z &\rightarrow z + \varphi_2 \cdot \mathrm{MLP}\left(\gamma_2 \cdot \mathrm{LN}(z) + \delta_2\right),
\end{align}

Here, $z$ is the transformer's hidden state, $\mathrm{LA}$ denotes local attention, $\mathrm{LN}$ represents layer normalisation, and $\mathrm{MLP}$ is a two-layer perceptron with a GELU \cite{gelu} activation between the layers.

\section{Kuramoto-Sivashinsky Equation}

As described in the main text, the emulators trained on the restrictive datasets shown in Fig. 2 are capable of both forecasting relaminarisation and initialising flows for the KS equation, as demonstrated in Fig. 6. In both cases, we observe transitions between states with $n$ kinks to $n+1$ and $n-1$ kinks and the emulators successfully reproduce these transitions, despite not encountering them during training.

Fig. 7(f) highlights the limitations of the Fourier Neural Operator (FNO) in generalising across domains of varying sizes. While the FNO generates realistic forecasts for domains of size $L=22$ and $L=144$ in Figures Fig. 7(d) and Fig. 7(e), it struggles to produce a plausible forecast when the domain size increases to $L=200$, as shown in Fig. 7(f) This limitation arises because the FNO explicitly models only the largest $N$ wavenumbers, relying on a $1 \times 1$ convolution to recover higher wavenumbers. For example, with $L=22$ and $N_x=56$ resolved grid points, only the highest $N=28$ wavenumbers are modelled by the FNO. When $L=200$, the system has approximately 31 unstable modes (estimated as $N_{\text{unstable}} \approx \frac{L}{2 \pi}$), which exceeds the capacity of an FNO trained on $L=22$ to capture all dynamically unstable modes at the larger domain size. In contrast, the local attention mechanism is designed to model local dynamics explicitly by learning a translation invariant stencil over the domain, allowing it to flexibly adapt to domains of arbitrary size without sacrificing the ability to resolve dynamics across scales.

\section{Beta-Plane Turbulence}

DNS of the two-layer beta-plane equation was performed using PYQG \cite{pyqg} in Python \cite{python} with the following parameters: $\beta = 5 \times 10^{-11} \, \text{s}^{-1}$, bottom layer damping $\mu = 4 \times 10^{-8} \, \text{s}^{-1}$, Rossby radius of deformation $R_d = 12.5 \, \text{km}$, depth of the upper layer $H_1 = 500 \text{m}$, $H_2 = 1500 \text{m}$ and the upper layer flow speed is $U_1 = 0.025 \text{m} \text{s}^{-1}$.

The small-scale dissipation (ssd) removes enstrophy cascading to small scales, preventing energy buildup at the grid scale that could cause numerical instability. Acting as a hyperviscosity-like term, it selectively damps high-wavenumber modes while preserving large-scale dynamics.  In Fourier space, this is applied via an exponential filter:  

\[
E_f = 
\begin{cases} 
\exp\left[C_{\text{ssd}}\,(\kappa^{\star} - \kappa_c)^4\right], & \quad \kappa \geq \kappa_c, \\ 
1, & \quad \text{otherwise}.
\end{cases}
\]

where $\kappa^{\star} = \sqrt{ (k\,\Delta x)^2 + (l\,\Delta y)^2 }$ is a non-dimensional wavenumber and \(\kappa_c\) is set to 65\% of the Nyquist scale, \((\kappa^{\star}_{\text{ny}} = \pi)\).  The constant \(C_{\text{ssd}}\) ensures energy at \(\kappa^{\star} = \pi\) vanishes within machine precision C$_{\text{ssd}} = \frac{\log 10^{-15}}{(0.35\, \pi)^4} \approx -23.5$.

\end{spacing}
\end{document}